\documentclass{article}
\usepackage{style/iclr2027_conference}

\usepackage{times}

\usepackage[T1]{fontenc}
\usepackage[utf8]{inputenc}
\usepackage{microtype}

\usepackage{amsmath}
\usepackage{amssymb}
\usepackage{amsfonts}
\usepackage{bm}

\usepackage{booktabs}
\usepackage{multirow}
\usepackage{array}
\usepackage{siunitx}
\usepackage{graphicx}
\graphicspath{{figures/}}
\usepackage{subcaption}
\usepackage[font=small,labelfont=bf]{caption}

\usepackage[ruled,vlined,linesnumbered]{algorithm2e}

\usepackage{url}
\usepackage{hyperref}
\hypersetup{
  colorlinks=true,
  linkcolor=black,
  citecolor=black,
  urlcolor=blue,
  breaklinks=true,
}
\usepackage[capitalise,noabbrev]{cleveref}

\usepackage{xcolor}
\newif\ifdraftmode
\draftmodefalse 

\ifdraftmode
  \newcommand{\todo}[1]{\textcolor{red}{[TODO: #1]}}
  \newcommand{\note}[1]{\textcolor{blue}{[#1]}}
\else
  \newcommand{\todo}[1]{}
  \newcommand{\note}[1]{}
\fi

\newcommand{\ci}[2]{$[#1,\,#2]$}

\newcommand{\zeroOf}[3]{$#1/#2$ (rule-of-three upper bound $\leq #3\%$)}

\newcommand{\exploratory}{\emph{exploratory}}

\newcommand{\clause}{C}
\newcommand{\spec}{S}
\newcommand{\specminus}[1]{\ensuremath{\spec \ominus #1}}

\newcommand{\AC}{\ensuremath{A_{\clause}}}
\newcommand{\ACdef}{\ensuremath{A_{\clause} = \Pr(\text{pass checker} \mid \spec)}}
\newcommand{\ACsemantics}{compliance as rendered in parseable code, a
  conservative lower bound on behavioral compliance}

\newcommand{\BC}{\ensuremath{B_{\clause}}}
\newcommand{\BCdef}{\ensuremath{B_{\clause} = A_{\clause} - \Pr(\text{pass} \mid \specminus{\clause})}}

\newcommand{\BCnameQualified}{incremental behavioral effect on a single-checker
  projection}

\newcommand{\horizon}{\ensuremath{h}}
\newcommand{\nmarkers}{\ensuremath{m}}

\newcommand{\sysname}{ReBIND}
\newcommand{\dataname}{RELAPSE-Code}

\newcommand{\armbare}{\textsc{bare}}
\newcommand{\armvrblind}{\textsc{vr-blind}}
\newcommand{\armvr}{\textsc{vr}}
\newcommand{\armrebind}{\textsc{rebind}}
\newcommand{\armtomb}{\textsc{bare+tomb}}
\newcommand{\armpos}{\textsc{bare+pos}}
\newcommand{\armplacebo}{\textsc{bare+placebo}}
\newcommand{\armfixed}{\textsc{fixed}}
\newcommand{\armadaptive}{\textsc{adaptive}}

\newcommand{\stadopted}{\textsf{adopted}}
\newcommand{\stredundant}{\textsf{redundant}}
\newcommand{\stunderpowered}{\textsf{underpowered}}
\newcommand{\stinert}{\textsf{inert}}
\newcommand{\stadverse}{\textsf{adverse}}
\newcommand{\stunconfident}{\texttt{confident=False}}

\newcommand{\clstrue}{\texttt{true\_relapse}}
\newcommand{\clsnear}{\texttt{near\_compliant}}
\newcommand{\clsoverlap}{\texttt{high\_overlap\_replacement}}
\newcommand{\clslegit}{\texttt{legit\_reference}}

\newcommand{\Lone}{L1}    
\newcommand{\Ltwo}{L2}    
\newcommand{\Lthree}{L3}  
\newcommand{\Lfive}{L5}   
\newcommand{\Lseven}{L7}  

\newcommand{\descriptive}{(descriptive)}

\newcommand{\checkerinvariance}{interventions modify only \texttt{surface\_text}
  and compilation parameters; \texttt{text} and \texttt{checker\_id} are never
  altered}

\newcommand{\NTasks}{67}                    
\newcommand{\NClauses}{201}                 
\newcommand{\NSlicesMain}{201}              
\newcommand{\NSlicesRevoked}{134}           
\newcommand{\NMarkersMain}{5}
\newcommand{\NMarkersVariants}{2 and 8}
\newcommand{\NBoot}{$10^{4}$}
\newcommand{\MaxTokens}{1024}
\newcommand{\CapTokens}{4000}
\newcommand{\MaxAttempts}{3}

\newcommand{\LeakMain}{0/201}
\newcommand{\LeakVariants}{0/201}           
\newcommand{\LeakVOne}{10/60}               

\newcommand{\CheckerVerified}{201/201}

\newcommand{\SpotCheckResult}{70/70}

\newcommand{\ProbeKMin}{3}
\newcommand{\ProbeKMax}{30}
\newcommand{\ProbeThreshold}{0.8}
\newcommand{\ProbeConf}{0.8}
\newcommand{\ProbeZeroBand}{0.2}
\newcommand{\ProbeBiasBound}{0.16}          
\newcommand{\ProbeWorstAcc}{0.712}
\newcommand{\ProbeWorstUncertain}{0.34}
\newcommand{\ProbeMeanSamplesSim}{14.1}     
\newcommand{\ProbeMeanSamplesObs}{21.2}     
\newcommand{\LadderMinGain}{0.2}
\newcommand{\FinalCheckCapacity}{3}
\newcommand{\KDiag}{10}

\newcommand{\EZeroPassAtOne}{0.821}
\newcommand{\EZeroPassAtOneCI}{\ci{0.746}{0.891}}
\newcommand{\EZeroReps}{3}
\newcommand{\InForceCompliance}{0.983}

\newcommand{\ScaleDelayedMTwo}{0.011}
\newcommand{\ScaleDelayedMFive}{0.238}
\newcommand{\ScaleDelayedMEight}{0.403}
\newcommand{\ScaleImmediateMTwo}{0.007}
\newcommand{\ScaleImmediateMFive}{0.105}
\newcommand{\ScaleImmediateMEight}{0.304}
\newcommand{\ScaleDiff}{$+0.392$}
\newcommand{\ScaleDiffCI}{\ci{+0.300}{+0.483}}
\newcommand{\ScaleDiffSigns}{$+45/-1$}
\newcommand{\ScaleDiffP}{$1.3\times10^{-12}$}
\newcommand{\RebindZeroScaling}{\zeroOf{0}{2968}{0.10}}
\newcommand{\MaxZeroScaling}{\zeroOf{0}{600}{0.5}}

\newcommand{\KimiCleanZero}{\zeroOf{0}{594}{0.51}}
\newcommand{\KimiCleanN}{200, 195, and 199}            
\newcommand{\KimiCleanCellUBs}{1.50\%, 1.54\%, and 1.51\%}
\newcommand{\KimiCleanEmptyRates}{0.5\%, 2.0\%, and 1.0\%}
\newcommand{\KimiMaxTokensRescue}{4096}
\newcommand{\KimiEmptyBareOld}{13--29\%}
\newcommand{\KimiEmptyRebindOld}{51--94\%}
\newcommand{\KimiEmptyRebindRescue}{20\%}              
\newcommand{\KimiPilotBareBefore}{22.9\%}
\newcommand{\KimiPilotBareAfter}{2.0\%}
\newcommand{\KimiPilotRebindBefore}{90.0\%}
\newcommand{\CompileParseFailScaling}{6.7--10.6\%}

\newcommand{\PrevalenceAtRevocation}{0.233}
\newcommand{\SurvivalAtZeroPlus}{0.767}
\newcommand{\SurvivalAtZeroPlusCI}{\ci{0.697}{0.832}}
\newcommand{\DepthDiff}{$-0.017$}             
\newcommand{\DepthDiffCI}{\ci{-0.075}{+0.036}}
\newcommand{\PrognosticDiff}{$+0.217$}
\newcommand{\PrognosticDiffCI}{\ci{+0.109}{+0.332}}

\newcommand{\SurvivalParseFail}{3/2010}
\newcommand{\SurvivalPfMaxShift}{0.05}     

\newcommand{\MappingPrecision}{1.000}
\newcommand{\MappingRecall}{1.000}
\newcommand{\AurocPrimary}{0.897}
\newcommand{\AurocPrimaryCI}{\ci{0.829}{0.963}}
\newcommand{\AurocHTwo}{0.915}
\newcommand{\AurocHFive}{0.831}
\newcommand{\AurocHEight}{0.897}
\newcommand{\AurocThreshold}{0.70}

\newcommand{\DetectorPrecision}{0.951}
\newcommand{\DetectorPrecisionCorrected}{1.000}
\newcommand{\DetectorLegitFPR}{0.020}
\newcommand{\DetectorRecall}{1.000}

\newcommand{\DetectorNBlind}{182}
\newcommand{\DetectorStrataPool}{1206}
\newcommand{\DetectorStrataFlagged}{82}
\newcommand{\DetectorStrataLegit}{60}
\newcommand{\DetectorStrataNone}{40}
\newcommand{\DetectorDisagreements}{15}    
\newcommand{\DetectorCRateOne}{0.297}      
\newcommand{\DetectorCRateTwo}{0.220}      
\newcommand{\DetectorInterAgreement}{0.918}
\newcommand{\DetectorInterKappaThree}{0.873}           
\newcommand{\DetectorInterKappaTwo}{0.984}             
\newcommand{\DetectorAdjPrecision}{1.000}
\newcommand{\DetectorAdjFPR}{0.000}
\newcommand{\EThreeAgreement}{0.954}
\newcommand{\EThreeKappa}{0.907}
\newcommand{\EThreeN}{762}
\newcommand{\EThreeParseableAgreement}{713/713}
\newcommand{\EThreeFormatUnstable}{49}
\newcommand{\EThreeFormatUnstablePct}{6.4\%}
\newcommand{\EThreeDiagAgreement}{0.914}
\newcommand{\EThreeDiagN}{32/35}
\newcommand{\EThreeDiagCount}{36}          
\newcommand{\BridgeKappa}{0.000}
\newcommand{\BridgeAgreement}{0.6}
\newcommand{\BridgeN}{10}
\newcommand{\EThreeSecondKappa}{1.000}
\newcommand{\EThreeSecondAgreement}{762/762}
\newcommand{\NullNClauses}{60}
\newcommand{\NullPNinetyFive}{0.200}
\newcommand{\NullMax}{0.300}
\newcommand{\NullMean}{$+0.007$}
\newcommand{\NullTheorySigma}{0.208}                   

\newcommand{\RestoreVrBlind}{0.250}
\newcommand{\RestoreVr}{0.025}
\newcommand{\RestoreRebind}{0.000}
\newcommand{\RestoreDiff}{0.192}
\newcommand{\RestoreDiffCI}{\ci{0.134}{0.251}}
\newcommand{\RestoreDiffP}{$\approx\!10^{-10}$}
\newcommand{\RestoreEpisodes}{402}
\newcommand{\CompileExtraGain}{$+0.0124$}
\newcommand{\CompileExtraGainCI}{\ci{+0.0025}{+0.0224}}
\newcommand{\RestoreTokenCost}{$+398$}
\newcommand{\RestoreScoreDiffCI}{\ci{-0.029}{+0.001}}

\newcommand{\TombBare}{0.135}
\newcommand{\TombNeutral}{0.087}
\newcommand{\TombPositive}{0.047}
\newcommand{\TombRebind}{0.000}
\newcommand{\TombRebindZero}{\zeroOf{0}{378}{0.8}}
\newcommand{\TombPrimary}{$+0.048$}           
\newcommand{\TombPrimaryCI}{\ci{+0.013}{+0.085}}
\newcommand{\TombPrimaryP}{0.041}
\newcommand{\PlaceboRate}{0.140}
\newcommand{\PlaceboTombDiff}{$+0.053$}       
\newcommand{\PlaceboTombDiffCI}{\ci{+0.018}{+0.090}}
\newcommand{\PlaceboTombP}{0.023}
\newcommand{\BarePlaceboDiff}{$-0.003$}       
\newcommand{\BarePlaceboDiffCI}{\ci{-0.045}{+0.040}}
\newcommand{\PlaceboDriftCI}{\ci{-0.0226}{+0.0226}}    
\newcommand{\TombFamilyBonfLBs}{$+0.0075$ and $+0.0125$} 
\newcommand{\TombFamilyCILevel}{97.5\%}     
\newcommand{\TombSignCorrectedP}{0.083}     
\newcommand{\PlaceboScoreUB}{1.3pp}         
\newcommand{\PlaceboLengthChars}{44}        
\newcommand{\TombSecondary}{$+0.040$}         
\newcommand{\TombSecondaryCI}{\ci{+0.010}{+0.075}}
\newcommand{\TombScoreHarm}{$-0.034$}
\newcommand{\TombScoreHarmCI}{\ci{-0.044}{-0.024}}
\newcommand{\TombWorstCase}{$+0.075$}
\newcommand{\TombWorstCaseCI}{\ci{+0.020}{+0.133}}
\newcommand{\TombParseFailCompile}{6.0\%}
\newcommand{\TombParseFailBare}{0.2\%}

\newcommand{\TombScoreNeutral}{$-0.005$}
\newcommand{\TombScoreNeutralCI}{\ci{-0.012}{+0.001}}
\newcommand{\TombScorePositive}{$-0.003$}
\newcommand{\TombScorePositiveCI}{\ci{-0.008}{+0.002}}

\newcommand{\LadderVr}{0.910}
\newcommand{\LadderFixed}{0.900}
\newcommand{\LadderAdaptive}{0.893}
\newcommand{\LadderDiff}{$-1.7$pp}            
\newcommand{\LadderDiffCI}{\ci{-5.0}{+1.2}}
\newcommand{\LadderDiffP}{0.48}
\newcommand{\LadderExcludedGain}{1.3pp}
\newcommand{\LadderUBIntervention}{+2.2pp}
\newcommand{\LadderUBAdaptivity}{+1.5pp}
\newcommand{\LadderResidualFailure}{9.0\%}
\newcommand{\LadderParseFailVr}{3.0\%}
\newcommand{\LadderParseFailFixed}{1.2\%}
\newcommand{\LadderParseFailAdaptive}{1.0\%}
\newcommand{\LadderParseFailP}{0.07}        
\newcommand{\LadderAdaptiveLOne}{122}
\newcommand{\LadderAdaptiveLFive}{27}
\newcommand{\LadderFixedLOne}{116}

\newcommand{\CostDeliveryFactor}{1.49$\times$}
\newcommand{\CostDeliveryTokensRebind}{38{,}979}
\newcommand{\CostDeliveryTokensBare}{26{,}198}
\newcommand{\CostOperatingFactor}{1.31$\times$}       
\newcommand{\CostClosedLoopFactor}{$\sim$15$\times$}

\newcommand{\CostBudgetCriterion}{5$\times$}
\newcommand{\CostMain}{$\approx$\$14}
\newcommand{\CostRescue}{$\approx$\$4}
\newcommand{\CostTotal}{\$17.87}
\newcommand{\CostTotalHedged}{under \$20}   
\newcommand{\CostQwenEightB}{\$5.58}
\newcommand{\CostQwenMax}{\$3.97}
\newcommand{\CostKimi}{\$8.31}
\newcommand{\PriceEightB}{0.1/0.4}
\newcommand{\PriceMax}{1.2/6.0}
\newcommand{\PriceKimi}{0.95/4.0}
\newcommand{\CostRescuePilot}{\$0.82}
\newcommand{\CostRescueFull}{\$2.79}
\newcommand{\CRNTokenExample}{1{,}032 versus 1{,}023}

\newcommand{\PilotRelapseRange}{0.10--0.65}

\iclrfinalcopy

\title{Dead Text or Binding Clause?\\
       Measuring and Restoring Constraint\\
       Influence in Black-Box LLM Dialogues}

\author{Haoyuan Zhu\\
Department of Electronic and Electrical Engineering\\
University of Sheffield\\
\texttt{hzhu51@sheffield.ac.uk}}

\begin{document}

\maketitle


\begin{abstract}
Multi-turn dialogues let users revoke constraints as easily as impose them,
but revocation does not reliably take effect: models keep enacting withdrawn requirements (occasionally beneath comments asserting their removal), a failure we call \emph{behavioral relapse}, or revocation inertia. No
existing instrument measures this influence per clause, predicts it before
delivery, or repairs it under matched budgets. \sysname{} closes the three
gaps through the model API alone: a contract ledger pairs every constraint
with an executable checker, records revocations as tombstones, and compiles
the net constraint state ahead of time into a single specification; a
sequential ablation probe measures per-clause adherence and incremental
behavioral effect; a repair ladder operates under token- and
attempt-matched budgets. On \dataname{} (\NTasks{} HumanEval tasks,
\NClauses{} verified checkers), relapse at an 8B operating point climbs
from \ScaleDelayedMTwo{} to \ScaleDelayedMEight{} as constraint load grows,
while stronger models sit at floor. Under matched checkers, model, and
budget, ahead-of-time compilation significantly reduces relapse against a
no-ledger verifier-retry baseline (\RestoreDiff{}, 95\% CI
\RestoreDiffCI{}, $p$ \RestoreDiffP{}); adaptive ladder interventions
stacked on top add no detectable gain (95\% confidence excludes gains
$\geq$ \LadderExcludedGain{}). The probe predicts relapse before delivery
(AUROC \AurocPrimary{}); a one-sentence tombstone note recovers about a
third of the compilation effect and survives a placebo control. At
\CostDeliveryFactor{} delivery overhead and \CostTotalHedged{} of API
compute for every result, revocation failure becomes a measurable,
predictable, and repairable property of dialogue state rather than an
invisible one.
\end{abstract}


\section{Introduction}
\label{sec:intro}

Large language models increasingly work in multi-turn dialogues, where the
requirements on an artifact are not fixed in advance but negotiated as the
conversation unfolds: users add constraints, amend them, and withdraw them,
and instruction-tuned models are expected to track the result
\citep{ouyang2022training}. A growing literature documents how models degrade under this regime: earlier instructions lose force as turns accumulate,
multi-turn capability lags matched single-turn capability, and what a model
heeds depends on where material sits in its context \citep{laban2025lost,
liu2024lost, li2024measuring}. That line of work studies constraints that
should bind but no longer do: live instructions decaying into dead text. This paper studies the mirror image: constraints that should no longer bind but still do. Ask a coding assistant to define an extra helper function,
\texttt{audit\_log}, in every solution; let it comply for a few tasks; then
withdraw the requirement: several turns later the assistant is still defining
\texttt{audit\_log}, occasionally right next to a comment asserting that the
function was removed as requested. We call the phenomenon \emph{behavioral
relapse} of revoked constraints, or \emph{revocation inertia}.

Relapse is not forgetting run in reverse, and the two failure modes are not
symmetric. In a stress pilot, in-force compliance failures stayed near zero while
relapse of revoked clauses reached \PilotRelapseRange{} \descriptive{}. Three ingredients make the event precise. The revocation is
\emph{delayed}: unrelated turns separate adoption from withdrawal. The
constraint was \emph{previously adopted}: it demonstrably shaped earlier
answers. And relapse is measured in \emph{behavior}, on the parsed artifact rather than in text, so a model that merely mentions a withdrawn requirement, for
instance to explain its withdrawal, is not thereby relapsing. At an 8B
operating point relapse climbs steeply with constraint load while stronger
controls show none (\cref{sec:exp:scaling}).

Adjacent lines of work (multi-turn evaluation, constraint-following
benchmarks, response-level intervention, contracts and dialogue state)
each stop short of this event (\cref{sec:related}); three gaps therefore
keep the phenomenon invisible. \emph{No measurement}: no instrument isolates, clause by clause, whether
a requirement no longer in force still shapes behavior (search protocol
in \cref{app:literature}). \emph{No prediction}: nothing flags, before
delivery, which dialogues are at risk of relapsing. \emph{No budget-matched
restoration}: comparisons of dialogue-repair interventions seldom hold
checkers, model, and token budget fixed at once, which conflates mechanism
with spend.

We close the three gaps with \sysname{} (Rebinding Diagnostics for Black-box
LLMs), which operates through the model API alone. \sysname{} maintains a \emph{contract ledger}: every user
constraint becomes a clause with an executable checker and a binding history;
revoking a clause writes a \emph{tombstone} (the record survives, the obligation does not), and the net state of in-force clauses is compiled ahead
of time into a single specification (\cref{sec:method}). On this substrate the system measures each clause's adherence and its incremental behavioral effect on a single-checker projection, diagnoses clauses into a five-state triage with an explicit non-committal state, and restores bindings through a repair ladder under matched budgets.
Evaluation runs on \dataname{}, an evaluation slice built from \NTasks{}
HumanEval tasks \citep{chen2021evaluating} with \NClauses{} human-verified
executable checkers, in which adoption, delay, and revocation are explicitly
controlled (\cref{sec:benchmark}).

\paragraph{Contributions.}
\begin{itemize}
  \item \textbf{Clause-level measurement and prospective prediction}
    (\cref{sec:formulation,sec:method}). Black-box per-clause measurement
    of adherence \AC{} and incremental behavioral effect \BC{} on a
    single-checker projection, with a simulation-calibrated sequential
    stopping rule and a five-state triage; checker judgments match blind
    human gold standards, and the diagnosis-time signal predicts later
    relapse (\cref{sec:exp:validity}).
  \item \textbf{Equal-budget restoration with an informative null}
    (\cref{sec:exp:restoration}). Under matched checkers, model, and token
    budget, ahead-of-time compilation of the net constraint state
    significantly reduces relapse against a no-ledger verifier-retry
    baseline (\RestoreDiff{}, 95\% CI \RestoreDiffCI{}, $p$
    \RestoreDiffP{}), while stacking adaptive ladder interventions on top
    yields no additional detectable pass-rate gain (95\% confidence
    excluding gains $\geq$ \LadderExcludedGain{}); costs are reported
    beside gains throughout.
  \item \textbf{\dataname{}, a relapse detector, and counterfactual
    tombstones} (\cref{sec:benchmark,sec:exp:restoration}). A
    leakage-screened evaluation slice isolating relapse of
    adopted-then-revoked constraints, a blind-validated four-class
    attribution detector (\cref{sec:benchmark:detector}), and a
    placebo-controlled counterfactual: a one-sentence tombstone note
    causally reduces relapse.
  \item \textbf{Phenomenon characterization}
    (\cref{sec:exp:scaling,sec:exp:temporal}). Relapse scales with
    constraint load at the 8B tier while stronger controls sit at floor, a
    capability gradient rather than an independent replication; onset is
    immediate at revocation, with no detectable depth accumulation.
  \item \textbf{Reliability--cost frontier} (\cref{sec:exp:cost}). Delivery
    overhead is measured against a pre-registered budget criterion, and all
    artifacts and protocols are released.
\end{itemize}


\section{Related Work}
\label{sec:related}

\sysname{} sits at the intersection of several lines of work, none of which
measures the influence of revoked constraints.

\paragraph{Multi-turn degradation and instruction forgetting.}
A growing body of evaluations reports that models drift away from earlier
instructions as dialogues lengthen, and that multi-turn capability lags matched single-turn capability \citep{laban2025lost, zheng2023judging, liu2024lost, li2024measuring}. The failure studied in this line is decay of live
constraints: instructions that should bind but no longer do. Where that
line asks when live instructions become dead text, we ask when revoked text
remains a binding clause.

\paragraph{Constraint-following benchmarks.}
Single-turn instruction following is scored by verifiable programmatic
checks \citep{zhou2023instruction} and by graded constraint hierarchies
\citep{jiang2024followbench}; multi-turn extensions add instructions across
turns and languages \citep{he2024multi}. Closer to our setting, DriftBench measures a knows-but-violates (KBV) rate: models accurately restate constraints they simultaneously violate \citep{kruthof2026driftbench};
SEQUOR, MCJudgeBench, and One-Battle-After-Another evaluate adherence to
stated requirements under multi-turn protocols \citep{canaverde2026sequor,
lee2026mcjudgebench, jia2025battle}. All of these instruments score compliance with what is currently required. \dataname{} isolates the complementary event (whether behavior that is no longer required persists) and operationalizes a KBV-style dissociation onto executable per-clause
checkers, measured clause by clause and turn by turn.

\paragraph{Interventions and constrained generation.}
Structured-output modes constrain decoding to a schema
\citep{willard2023efficient}; prefill fixes the opening tokens of a
response; and format-restriction studies document a tension between enforced form and task performance \citep{tam2024speak, sclar2024quantifying}.
On the repair side, models revise outputs against
self- or execution feedback \citep{madaan2023selfrefine, shinn2023reflexion,
chen2024teaching}, with the headroom of self-repair itself under scrutiny
\citep{olausson2024selfrepair}. These mechanisms govern the form, or the
retry, of one response. Our problem is dialogue-state management, deciding which clauses are in force at all, and the two levels compose: structural
forcing enters our repair ladder as a single rung (\Lfive{}) and
verifier-retry as another (\Lseven{}), applied per diagnosis rather than
globally and compared under matched budgets.

\paragraph{Contracts, dialogue state, and verification.}
Design-by-contract treats obligations as first-class program objects
\citep{meyer1992applying}; contract-based evaluation of generated code
verifies artifacts against specifications, as in ContractEval's PACT
framework \citep{lim2025contracteval}. That line treats the specification as
fixed and asks whether an artifact satisfies it. Task-oriented dialogue
systems, conversely, have long tracked evolving user goals as mutable
belief state \citep{budzianowski2018multiwoz}, but track them as slot
values to fill rather than as obligations whose per-clause behavioral force
can be measured. \sysname{} manages the specification itself as mutable dialogue state (creation, revocation, replacement) and asks whether
withdrawn parts of it keep leaking into behavior.

\paragraph{Statistical methodology.}
The experimental designs are imported from biostatistics rather than
benchmark practice: negative-control reasoning motivates the placebo arms
\citep{lipsitch2010negative, ye2025role, guo2024estimating}; the probe's
stopping rule descends from sequential tests \citep{wald1945sequential};
and survival-style analyses of dialogue robustness \citep{li2025time}
motivate a temporal analysis kept deliberately cross-sectional, because a
relapsed clause can return to compliance.


\section{Problem Formulation and the \dataname{} Benchmark}
\label{sec:formulation}

We model a dialogue as a sequence of turns that create, revoke, and replace
constraints on a final artifact; evaluation runs on \dataname{}, an
evaluation slice in which relapse is decided by executable
per-clause checkers rather than by judges.

\subsection{Dialogues, clauses, and revocation}
\label{sec:formulation:clauses}

A clause records its normative content (\texttt{text}), the phrasing the
model actually sees (\texttt{surface\_text}), an executable verifier (a program, not a model judge) that decides on a given artifact whether the
clause is satisfied (\texttt{checker\_id}), its position, and its
diagnostic and intervention record (\texttt{binding\_history}). Revoking a clause writes a \emph{tombstone}: the clause leaves the in-force set while its record is retained, optionally with a positive replacement; the \emph{net state} at any turn is the set of in-force clauses, against which adherence, scoring, and every comparison below are evaluated.

A \emph{behavioral relapse} occurs when a later artifact satisfies the
checker of a tombstoned clause although no in-force clause requires that
behavior. Its three constitutive ingredients are as stated in \cref{sec:intro};
\cref{sec:benchmark} controls each explicitly.

\subsection{Adherence, incremental effect, and a five-state taxonomy}
\label{sec:formulation:taxonomy}

For a clause $\clause$ under specification $\spec$, adherence is \ACdef.
Artifacts whose extracted code does not parse count as non-compliant, so
\AC{} measures \ACsemantics{}; \cref{sec:exp:validity} quantifies exactly
where this conservative reading and human judgment part ways. The clause's
\BCnameQualified{} is \BCdef, where \specminus{\clause} replaces the clause's
surface text with a neutral placeholder of equal length at the same position
(\cref{sec:method:ablation}). \BC{} captures the clause's marginal effect on its own checker and makes no claim about the full generation distribution.

Crossing the two measurements yields the five-state triage of
\cref{tab:taxonomy}, which routes each clause to a system action; state
assignment is a posterior classification under the sequential stopping rule
of \cref{sec:method:probe}, with non-committal outcomes explicit rather than missing.

\begin{table}[t]
  \centering
  \caption{Five-state triage from the two clause-level measurements. States
    are posterior classifications from the sequential probe
    (\cref{sec:method:probe}): \AC{} is classified against threshold
    \ProbeThreshold{} and \BC{} against a zero band of $\pm$\ProbeZeroBand{},
    each side to posterior confidence \ProbeConf{}. A probe that stops
    without reaching confidence returns \stunconfident{}, an explicit
    non-committal state rather than a missing value; downstream components do
    not silently adopt such diagnoses. Actions reference the repair ladder of
    \cref{sec:method:ladder}.}
  \label{tab:taxonomy}
  \small
  \begin{tabular}{llll}
    \toprule
    \AC & \BC & State & System action \\
    \midrule
    high & $>0$       & \stadopted{}      & deliver \\
    high & $\approx0$ & \stredundant{}    & deliver; may deprioritize to save probe budget \\
    low  & $>0$       & \stunderpowered{} & strengthen the binding (\Lthree{}, \Lfive{}) \\
    low  & $\approx0$ & \stinert{}        & re-bind through another channel (\Lone{}, \Ltwo{}) \\
    any  & $<0$       & \stadverse{}      & roll back last rewrite; flag for attribution \\
    \bottomrule
  \end{tabular}
\end{table}


\subsection{Benchmark construction}
\label{sec:benchmark}

\paragraph{Gold tasks and checkers.}
We start from a gold subset of \NTasks{} HumanEval tasks
\citep{chen2021evaluating}. Each task carries
three executable checkers (its unit tests, a presence check on the entry point, and a standard-library-only check) for \NClauses{} clauses in total.
Clause-to-checker mappings were annotated by two independent annotators with
precision and recall of \MappingPrecision{}/\MappingRecall{}, and all
\CheckerVerified{} checkers were verified to accept a gold solution and to
reject seeded counterexamples (\cref{app:benchmark}).

\paragraph{Slices.}
Each task yields three dialogue scripts: \emph{immediate} revocation,
\emph{delayed} revocation (unrelated turns separate the marker block from
the revocation), and a no-revocation \emph{control}. A script injects
$\nmarkers$ marker clauses, each requiring one additional empty helper function, and, outside the control, revokes the second; the remaining
markers stay in force, so no in-force clause requires the revoked behavior.
The main set fixes $\nmarkers = \NMarkersMain$ (\NSlicesMain{} slices,
\NSlicesRevoked{} with a revocation); variant sets at loads of
\NMarkersVariants{} replicate these counts (marker-pool construction in
\cref{app:benchmark}). Scripts contain user turns only and each episode
elicits a single final implementation; adoption therefore holds in distribution rather than in-context: marker-form clauses show near-ceiling compliance whenever in force (\cref{sec:exp:setup}).

\paragraph{Leakage control and audit.}
Two screens guard against scripts leaking task solutions
\citep{jacovi2023stop}: a mechanical pre-screen and a live screen that must
fail to solve the task from auxiliary turns alone (definitions in
\cref{app:benchmark}). The released sets screen clean: \LeakMain{} on the main set and
\LeakVariants{} on each variant; a \SpotCheckResult{} manual audit found
no defects (\cref{app:benchmark}).

\subsection{Detector and release}
\label{sec:benchmark:detector}

\paragraph{Relapse detector and sub-evaluation.}
Final artifacts are parsed by concatenating markdown code fences; unparseable
output is scored as non-compliant, per the conservative semantics of
\cref{sec:formulation:taxonomy}. Detected marker behavior is attributed to
one of four classes, of which only \clstrue{} (the tombstoned clause's checker passes and no in-force clause requires the behavior) blocks delivery and enters relapse rates; the three non-blocking classes
(near-compliant, high-overlap replacement, legitimate reference) are defined
in \cref{app:benchmark}, and the detector carries its own blind human
sub-evaluation (\cref{sec:exp:validity}). The slice generator, both screens, a data card \citep{gebru2021datasheets},
and all annotation protocols accompany the paper
(\cref{app:benchmark,app:annotation}).


\section{\sysname: Ledger-Based Measurement and Restoration}
\label{sec:method}

\sysname{} treats a dialogue's constraints the way a medical chart treats
prescriptions: every requirement is a ledger entry, withdrawal leaves a
tombstone rather than an erasure, diagnosis proceeds by controlled
ablation, and repair escalates along a ladder with rollback
(\cref{fig:method}; \cref{sec:method:ledger}--\cref{sec:method:ladder} walk
through the pillars).

\subsection{Contract ledger and specification compilation}
\label{sec:method:ledger}

The ledger records every clause of \cref{sec:formulation:clauses} together
with its binding history; revocation tombstones a clause as described there.
Compilation maps the net state to a pseudo-single-turn specification: a task
header, the in-force clauses as a numbered list, and an optional
\emph{final-check block} of capacity \FinalCheckCapacity{} that restates
selected clauses at the end of the specification, the carrier of the
\Lone{} intervention (\cref{sec:method:ladder}). \Cref{app:examples} shows a
dialogue and its compiled form side by side; ahead-of-time compilation is
itself the \armrebind{} arm of \cref{sec:exp:restoration}. One discipline
governs all interventions, and it anchors every causal reading in this paper: \checkerinvariance.

\subsection{Science-mode ablation}
\label{sec:method:ablation}

To measure \BC{}, the probe compares the full specification with an ablated
copy in which the clause's surface text is replaced by a neutral placeholder
of the same length at the same position, so layout is held fixed and the
only difference is the clause's content. Deleting the clause instead is not
an interchangeable shortcut: a bridge comparison of the two ablation modes
on matched clauses agreed no better than chance ($\kappa = \BridgeKappa$,
raw agreement \BridgeAgreement{}, $n = \BridgeN$ clauses). All statistics in
this paper use this science-mode (equal-length neutral-placeholder)
ablation.

\subsection{Sequential probe}
\label{sec:method:probe}

Each probe round draws one paired sample under the full and the ablated
specification. The \AC{} side classifies the full-specification pass rate
against threshold \ProbeThreshold{} under a Beta--Binomial posterior; the
\BC{} side classifies the paired difference against a zero band of
$\pm$\ProbeZeroBand{} via a three-way posterior from the two arms' convolved
Beta posteriors. The probe
stops when both classifications reach posterior confidence \ProbeConf{}, and
otherwise at $k_{\max} = \ProbeKMax$ rounds ($k_{\min} = \ProbeKMin$). The stopping rule was calibrated by paired power simulation injected into
the production implementation, so simulated and deployed rule families
cannot drift apart; sequential inflation is absorbed into the simulated
operating characteristics (\cref{app:probe}). In deployment the probe averaged \ProbeMeanSamplesObs{} samples per diagnosis; an executed placebo-clause calibration corroborates the zero band (\cref{app:probe}), and the running estimate doubles as a prospective relapse signal (\cref{sec:exp:validity}).

\subsection{Repair ladder and equal-budget accounting}
\label{sec:method:ladder}

Diagnoses route to repairs along a ladder: \Lone{} promotes a clause into
the final-check block; \Ltwo{} rewrites its surface text imperatively;
\Lthree{} attaches a minimal contrastive example, verified against the
clause's own checker and withheld if that verification fails; \Lfive{}
forces structure by prefilling the opening of the response; \Lseven{}
retries against a pointed violation report, where the model sees the report
only, not its previous outputs; this rung is also the carrier of the
verifier-retry baseline of \cref{sec:exp:restoration}. (Rung numbers index a larger design
space; only these five are implemented.) Each rung is a client-side
transformation consuming no model call; retries draw on the same metered
budget as every arm (\CapTokens{} tokens, \MaxAttempts{} attempts per
episode; \cref{app:cost}). After an intervention the probe re-tests: the change is kept if adherence improves by at least \LadderMinGain{} or the clause confidently reaches \stadopted{} or \stredundant{}; regression elsewhere triggers rollback to the pre-intervention surface, and outcomes append to the clause's binding history.


\section{Experiments}
\label{sec:experiments}

\subsection{Setup}
\label{sec:exp:setup}

Three models are evaluated through their public APIs: qwen3-8b as the
operating point (the capability tier at which the stress pilot located the
phenomenon), qwen3-max as a same-family ceiling control
\citep{yang2025qwen3}, and kimi-k2.7-code as a cross-family control
\citep{kimiteam2025kimi}. Episodes use fixed decoding settings with a
per-episode output cap of \MaxTokens{} tokens; the cross-family rescue grid
raises the cap to \KimiMaxTokensRescue{} as a declared per-model deviation
(\cref{app:prereg}). Cross-arm common random numbers are best-effort
provider-side seed determinism, which widens intervals rather than biasing the paired contrasts.

Unless stated otherwise, intervals are percentile bootstrap 95\% CIs over
\NTasks{} task clusters (\NBoot{} resamples)
\citep{efron1994introduction, miller2024adding}; clustering is by task (same-task slices share prompts and markers, so finer clustering understates variance). Each experiment carries exactly one pre-registered confirmatory primary
test \citep{vanmiltenburg2021preregistering}; every other outcome is
reported with a nominal CI and labeled \exploratory{}. Sensitivity analyses use a cluster-level exact
sign test, a check on percentile-bootstrap coverage for rare binary
outcomes; multiplicity across clause families is controlled by
Benjamini--Hochberg \citep{benjamini1995controlling}. The pre-registration,
the decision-record chain, and all registered deviations appear in
\cref{app:prereg}. Contrasts between arms are paired at the episode level
and computed over episodes with a defined outcome in both arms; a contrast therefore need not equal the difference of the pooled arm rates reported beside it; arm names are set in small caps (\armbare{}, \armvr{}, \armrebind{}).

Two anchors calibrate what follows: unconstrained pass@1 at the operating
point is \EZeroPassAtOne{} \EZeroPassAtOneCI{} (\EZeroReps{} repetitions per
task), and compliance under \armbare{}, the unintervened arm that sees the dialogue script as-is (\cref{app:examples}), averaged over in-force clauses is \InForceCompliance{} (inflated by the
easy marker clauses, so not comparable to pass@1), the empirical basis for
the adoption-in-distribution claim of \cref{sec:benchmark}.

\subsection{Relapse scales with constraint load}
\label{sec:exp:scaling}

\begin{figure}[t]
  \centering
  \begin{subfigure}[t]{0.48\linewidth}
    \centering
    \includegraphics[width=\linewidth]{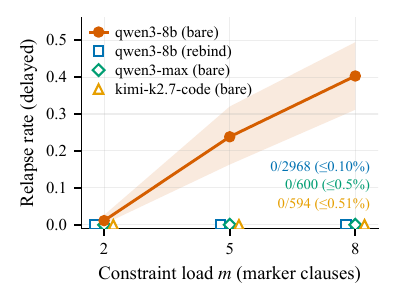}
    \caption{} 
    \label{fig:scaling}
  \end{subfigure}\hfill
  \begin{subfigure}[t]{0.48\linewidth}
    \centering
    \includegraphics[width=\linewidth]{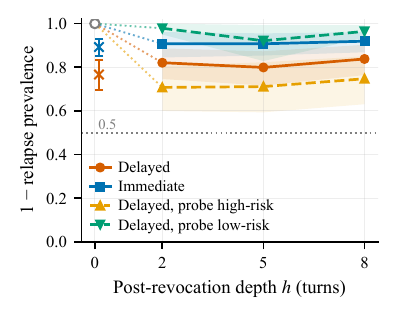}
    \caption{} 
    \label{fig:prevalence}
  \end{subfigure}
  \caption{\textbf{Behavioral relapse of revoked constraints: (a) load
    scaling; (b) temporal shape.} \textbf{(a)}~Relapse rate of the revoked
    marker clause under delayed revocation against injected constraint
    load $m$, for the operating-point model qwen3-8b under \armbare{} and
    \armrebind{} and for the ceiling and cross-family controls; shaded
    bands are 95\% cluster-bootstrap CIs over \NTasks{} task clusters, and
    zero-relapse series carry denominators and rule-of-three upper bounds
    as matching-color annotations in the panel. Pre-registered primary contrast ($m{=}8$ vs $m{=}2$):
    \ScaleDiff{} \ScaleDiffCI{} ($p =$ \ScaleDiffP{}). Results under
    compilation, the format-instability cost, and the unusable kimi
    compiled cell (not shown) are reported in \cref{sec:exp:scaling};
    per-cell figures are in \cref{app:results}.
    \textbf{(b)}~One minus relapse prevalence against post-revocation
    depth $\horizon$ (cross-sectional current-status reading; four strata
    with 95\% bands; anchor at $(0,1)$ by convention, drawn with an open
    marker). Crosses mark directly sampled at-revocation references; the
    probe-risk stratification shares its source with the prospective AUROC
    of \cref{sec:exp:validity}; the dotted line marks first crossing of
    one half, not reached within the observed horizon
    (\cref{sec:exp:temporal}).}
  \label{fig:phenomenon}
\end{figure}

Relapse grows steeply with constraint load at the operating point and sits
at floor everywhere else (\cref{fig:scaling}). The pre-registered primary
test compares delayed-revocation relapse at loads $m{=}8$ and $m{=}2$:
the difference is \ScaleDiff{} \ScaleDiffCI{} (cluster sign test
\ScaleDiffSigns{}, $p =$ \ScaleDiffP{}), over grid points
\ScaleDelayedMTwo{}, \ScaleDelayedMFive{}, and \ScaleDelayedMEight{}. Immediate-revocation rates are monotone and smaller in magnitude
\descriptive{} (full table in \cref{app:results}).

Under ledger compilation the picture inverts: observed relapse across all
twelve \armrebind{} cells is \RebindZeroScaling{}
\citep{hanley1983nothing}. The cost is reported
beside the gain: compilation destabilizes output format in
\CompileParseFailScaling{} of episodes (unparseable artifacts are excluded
from relapse denominators and scored as non-compliant elsewhere). The ceiling control shows no observed events (\MaxZeroScaling{}), and the cross-family control on the rescue grid likewise none (\KimiCleanZero{}, descriptive; per-cell figures in \cref{app:results}); its compiled cell remains unusable and is reported as
such rather than imputed, with the superseded grid and its
empty-completion artifact documented in \cref{app:results}. Across models this is a capability gradient (the controls sit at floor, so cross-model transfer correlations are undefined), not an independent replication.

\subsection{Temporal shape: immediate onset without depth accumulation}
\label{sec:exp:temporal}

Relapse propensity is already present at the moment of revocation and shows
no detectable accumulation with dialogue depth (\cref{fig:prevalence}). The
at-revocation reference is directly sampled: prevalence
\PrevalenceAtRevocation{} at the revocation turn, i.e.\ a non-relapse
prevalence of \SurvivalAtZeroPlus{} \SurvivalAtZeroPlusCI{}; the paired
difference between the deepest and shallowest observed horizons is
\DepthDiff{} \DepthDiffCI{}; and the depth at which prevalence would first
cross one half lies beyond the observed horizon in every bootstrap
replicate, reported as right-censored rather than extrapolated. Stratifying
by the probe's diagnosis-time risk signal separates the curves by
\PrognosticDiff{} \PrognosticDiffCI{} at the primary horizon, another view of the same signal as the AUROC of \cref{sec:exp:validity} and not an independent replication. Conventions (the $(0,1)$ anchor) and sensitivity (\SurvivalParseFail{}
unparseable episodes) are in \cref{app:results}.

\subsection{Measurement validity}
\label{sec:exp:validity}

The checker stack agrees with blind human judgment on every parseable
sample, and its divergences concentrate at the format boundary, in one direction. On \EThreeN{} blind-annotated samples, checker-human agreement
is \EThreeAgreement{} ($\kappa = \EThreeKappa$)
\citep{cohen1960coefficient}; on the \EThreeParseableAgreement{} parseable
samples there is no disagreement at all. The remaining \EThreeFormatUnstable{} samples
(\EThreeFormatUnstablePct{}) are format-unstable (correct-looking code wrapped in unparseable markdown), where the checker scores non-compliance
by its conservative semantics while human raters often judge the behavior
compliant. An
independent second annotator re-rated the full set with complete
concordance (\EThreeSecondAgreement{}, $\kappa = \EThreeSecondKappa$).

The relapse detector was audited against a blind, stratified human gold
standard of \DetectorNBlind{} artifacts: precision \DetectorPrecision{},
legitimate-reference false-positive rate \DetectorLegitFPR{}, recall
\DetectorRecall{} within the audited sample. Each of the handful of disagreements proved a human miss, so corrected
precision reads
\DetectorPrecisionCorrected{} (adjudication caveats in
\cref{app:annotation}). A second annotator rated the same set independently
with agreement \DetectorInterAgreement{} (dual-annotation protocol and
adjudicated figures in \cref{app:annotation}).

Prospectively, the probe's diagnosis-time signal predicts relapse at the
pre-registered horizon with AUROC \AurocPrimary{} \AurocPrimaryCI{},
against an adequacy criterion of \AurocThreshold{}; across horizons the
curve reads \AurocHTwo{}, \AurocHFive{}, and \AurocHEight{}
(\cref{fig:diagnosis}), staying above the criterion at every distance with
no monotone trend.

\subsection{Restoration under a matched budget}
\label{sec:exp:restoration}

\begin{table}[t]
  \centering
  \caption{The restoration spectrum under matched budgets. Arm rows give
    point estimates; each block closes with its pre-registered confirmatory
    contrast (95\% cluster-bootstrap CI over \NTasks{} task clusters; the
    tombstone family declares two confirmatory contrasts and adjusts by
    Bonferroni). Contrasts are episode-paired (\cref{sec:exp:setup}) and
    need not equal differences of the arm rates above them. Zeros carry
    their denominators: the compiled arm of the tombstone block is
    \TombRebindZero{}. Cost columns are part of the result: token spend
    and format instability sit beside the gains, and score costs are
    reported alongside them in \cref{sec:exp:restoration}.}
  \label{tab:restoration}
  \footnotesize
  \renewcommand{\arraystretch}{0.95}
  \begin{tabular}{lll}
    \toprule
    Arm & Outcome & Cost \\
    \midrule
    \multicolumn{3}{l}{\emph{Equal-budget restoration} (relapse rate;
      \RestoreEpisodes{} episodes per arm)} \\
    \armvrblind{} & \RestoreVrBlind{} & -- \\
    \armvr{}      & \RestoreVr{}      & -- \\
    \armrebind{}  & \RestoreRebind{}  & \RestoreTokenCost{} tokens \\
    \multicolumn{3}{l}{\quad primary: \armvrblind{} $-$ \armrebind{} $=$
      \RestoreDiff{} \RestoreDiffCI{}, $p$ \RestoreDiffP{}} \\
    \midrule
    \multicolumn{3}{l}{\emph{Tombstone counterfactual} (relapse rate,
      parseable samples; one completion per episode)} \\
    \armbare{}    & \TombBare{}     & parse-fail \TombParseFailBare{} \\
    \armplacebo{} & \PlaceboRate{}  & -- \\
    \armtomb{}    & \TombNeutral{}  & -- \\
    \armpos{}     & \TombPositive{} & -- \\
    \armrebind{}  & \TombRebind{}   & parse-fail \TombParseFailCompile{} \\
    \multicolumn{3}{l}{\quad confirmatory:
      \armbare{} $-$ \armtomb{} $=$ \TombPrimary{} \TombPrimaryCI{},
      $p = \TombPrimaryP{}$;} \\
    \multicolumn{3}{l}{\quad \phantom{confirmatory:}
      \armplacebo{} $-$ \armtomb{} $=$ \PlaceboTombDiff{}
      \PlaceboTombDiffCI{}, $p = \PlaceboTombP{}$} \\
    \multicolumn{3}{l}{\quad \exploratory{}: \armtomb{} $-$ \armpos{} $=$
      \TombSecondary{} \TombSecondaryCI{};} \\
    \multicolumn{3}{l}{\quad \phantom{\exploratory{}:}
      \armbare{} $-$ \armplacebo{} $=$ \BarePlaceboDiff{}
      \BarePlaceboDiffCI{}} \\
    \midrule
    \multicolumn{3}{l}{\emph{Adaptive ladder} (final pass rate; compiled
      form and budget held fixed)} \\
    \armvr{}       & \LadderVr{}       & parse-fail \LadderParseFailVr{} \\
    \armfixed{}    & \LadderFixed{}    & parse-fail \LadderParseFailFixed{} \\
    \armadaptive{} & \LadderAdaptive{} & parse-fail \LadderParseFailAdaptive{} \\
    \multicolumn{3}{l}{\quad primary: \armadaptive{} $-$ \armvr{} $=$
      \LadderDiff{} \LadderDiffCI{}, $p = \LadderDiffP{}$; excludes gains
      $\geq$ \LadderExcludedGain{}} \\
    \bottomrule
  \end{tabular}
\end{table}

Under matched checkers, model, and token budget, the ledger's value
concentrates in making relapse detectable and removable ahead of time
(\cref{tab:restoration}). The three-arm comparison fixes the intervention
budget and varies only ledger access: \armvrblind{} retries against a
verifier that cannot see revocation state, so relapse is undetectable to
it; \armvr{} feeds the ledger's violation report back after generation;
\armrebind{} compiles the net state ahead of time. Relapse rates read
\RestoreVrBlind{}, \RestoreVr{}, and \RestoreRebind{} respectively. The
pre-registered primary test confirms that compilation significantly reduces
relapse against the no-ledger baseline (\RestoreDiff{} \RestoreDiffCI{},
$p$ \RestoreDiffP{}; \RestoreEpisodes{} episodes, \NTasks{} clusters). Decomposing the mechanism, most of the value lies in
detectability itself (\RestoreVrBlind{} $\to$ \RestoreVr{}); compiling
ahead of time adds a further \CompileExtraGain{} \CompileExtraGainCI{}. The
costs sit beside the gains: the compiled arm spends \RestoreTokenCost{}
tokens per task, and its score difference is \RestoreScoreDiffCI{}, an interval containing zero.

The tombstone counterfactual isolates what the revocation record alone
contributes (\cref{tab:restoration}; arm-level view in
\cref{fig:tombstone}). Relapse falls monotonically \descriptive{} across
\armbare{} (\TombBare{}), \armplacebo{} (\PlaceboRate{}), \armtomb{}
(\TombNeutral{}), \armpos{} (\TombPositive{}), and \armrebind{}
(\TombRebind{}; \TombRebindZero{}). The family's two confirmatory contrasts
both survive Bonferroni adjustment (\TombFamilyCILevel{} CI lower bounds
\TombFamilyBonfLBs{}). A one-sentence tombstone note reduces relapse by
\TombPrimary{} \TombPrimaryCI{} ($p = \TombPrimaryP{}$; the adjusted
cluster sign test reads \TombSignCorrectedP{}, a borderline value we report
as such). The same note beats a near-equal-length irrelevant note by
\PlaceboTombDiff{} \PlaceboTombDiffCI{} ($p = \PlaceboTombP{}$), tying the
effect to revocation semantics rather than to appended text as such. The
\armbare{}--\armplacebo{} comparison (\cref{tab:restoration},
\exploratory{}) supports only the absence of a detected note-per-se effect
at the observed scale, not its exclusion (the placebo arm was collected in
a separate run and paired by slice and repetition; drift check in
\cref{app:results}). Secondary orderings are \exploratory{}
(\cref{tab:restoration}). Costs, again beside gains: the note costs at most
\PlaceboScoreUB{} in score (nominal), while full compilation costs
\TombScoreHarm{} \TombScoreHarmCI{} in score and raises format instability
to \TombParseFailCompile{} against \TombParseFailBare{} for \armbare{}; a
worst-case bound treating every unparseable compiled episode as relapse
does not flip the direction (\cref{app:results}). The practical spectrum:
one-sentence notes recover a third (neutral tombstone) to two-thirds
(positive replacement) of the effect at near-zero cost; compilation removes
the remainder and pays measurable costs.

Stacking adaptive routing on top of this machinery adds nothing detectable.
The pre-registered three-arm comparison holds compiled form and budget
fixed: \armvr{} (final pass rate \LadderVr{}), \armfixed{} (a constant
\Lone{} intervention, \LadderFixed{}), and \armadaptive{} (violation-typed
routing, \LadderAdaptive{}). The primary test reads \LadderDiff{}
\LadderDiffCI{} ($p = \LadderDiffP{}$), with 95\% confidence excluding
gains of \LadderExcludedGain{} or more. Mechanism upper bounds are in \cref{app:results}; interpretation of the
headroom is deferred to \cref{sec:discussion}. Secondary outcomes are in \cref{app:results}. All three arms show zero observed relapse, which
re-states the compiled-form result rather than re-verifying it, as the
design carries no \armbare{} arm.

\subsection{Cost}
\label{sec:exp:cost}

The reliability gains price out at a delivery overhead of
\CostDeliveryFactor{} against the pre-registered \CostBudgetCriterion{} criterion (\CostDeliveryTokensRebind{} tokens per delivered task under \armrebind{} versus \CostDeliveryTokensBare{} under \armbare{}), and \CostOperatingFactor{} at the operating-point measurement. Closing the loop with sequential probing costs \CostClosedLoopFactor{}
per task un-amortized.
Total API compute for every result in this paper is \CostTotal{} at
billed-confirmed unit prices (\CostMain{} main experiments, \CostRescue{}
cross-family rescue; full accounting and per-model totals in
\cref{app:cost}, \cref{tab:cost}).


\section{Discussion and Limitations}
\label{sec:discussion}

\paragraph{Why the ladder adds nothing at this scale.}
The pre-registered null of \cref{sec:exp:restoration} is informative about
the baseline, not only about the ladder: with compiled form and matched
budgets, verifier-retry already leaves residual failure at
\LadderResidualFailure{}, so the room in which routing could show value is
nearly gone; at this headroom, ``interventions do not help here'' and
``routing adds nothing over a fixed intervention'' are observationally
indistinguishable. Whether routing has value at harder operating points, where residual
failure is substantial, is an open question rather than an expectation.

\paragraph{Compliance and form are decoupled.}
The two audits of \cref{sec:exp:validity} expose one decoupling in mirror
image. Models relapse while asserting compliance in adjacent comments, so self-reports about constraint state cannot substitute for behavioral checking. Conversely, models comply while failing formally, and this
accounts for every checker--human divergence, in the strict direction
(\cref{sec:exp:validity}).

\paragraph{What the tombstone counterfactual isolates.}
A single tombstone note carries about half of what the strongest
one-sentence intervention delivers (\cref{tab:restoration}), and the
placebo arm ties the effect to revocation semantics
rather than to appended text as such, with two boundaries stated plainly:
the placebo is a single frozen text at a single position, and a note-per-se
effect is only bounded, not excluded (\cref{sec:exp:restoration}). Full
compilation removes the remainder and pays for it in score and format
stability (\cref{tab:restoration}); which point on this spectrum to operate
at is a deployment decision, not a fixed recommendation.

\paragraph{Limitations.}
(i)~one task domain (Python code), Chinese-language scripts, an 8B
operating point with stronger models only as controls; (ii)~\AC{} is
\ACsemantics{}; (iii)~probe-driven routing and bandit selection remain
unverified (demoted to open questions by the null); (iv)~the placebo
control is one frozen text at one position; (v)~the temporal analysis is
cross-sectional, with no extrapolation beyond the observed horizon;
(vi)~the no-ledger baseline is operationalized conservatively, which if
anything flatters it; (vii)~common random numbers are best-effort
provider-side seed determinism, not exact replay; (viii)~the cross-family
control runs under a declared per-model output-cap deviation with its compiled cell unusable, a capability gradient rather than an independent replication; (ix)~the prognostic stratification shares its source with the
AUROC; (x)~three procedural obligations open at pre-registration were closed
before submission (\cref{app:prereg}).


\section{Conclusion}
\label{sec:conclusion}

Revoked constraints can remain behaviorally binding: at an 8B operating
point, relapse of withdrawn requirements scales steeply with constraint
load, appears at the moment of revocation, and does not need dialogue depth
to accumulate. Maintaining the dialogue's net constraint state in a contract ledger, and compiling it ahead of time, removes the observed relapse
under matched checkers, model, and token budget, while a placebo-controlled
counterfactual shows that even the one-sentence tombstone note carries
real weight; adaptive intervention routing on top adds nothing detectable.
These results are bounded by their setting (one code domain, one operating-point capability tier, horizons up to eight turns), and the mechanisms by which stronger models hold their floor remain unmeasured;
induction-head-style in-context copying is one candidate substrate for
white-box follow-up \citep{olsson2022context}.


\section*{Ethics Statement}
This work evaluates publicly served language models on programming tasks
derived from HumanEval; no personal data, human subjects beyond the
annotators, or sensitive content are involved. Two annotators produced the
gold standards: rater one is an author, rater two an independent external
annotator; both worked blind to system judgments through self-contained
annotation interfaces, and disagreements were adjudicated under a rule
frozen before annotation began (\cref{app:annotation}). The measurement
techniques here audit whether systems still enact withdrawn instructions;
we see primarily defensive uses (verifying that revocations take effect) and no capability uplift beyond what the underlying APIs already provide.

\section*{Reproducibility Statement}
Every number in this paper originates from a registered run and is cited
through a single macro source checked against the run reports. The release
accompanying the paper contains the slice generator, marker name pool, both
leakage screens, checker definitions, annotation interfaces and protocols,
the pre-registration with its decision-record chain and all registered
deviations (\cref{app:prereg}), frozen probe and intervention parameters
(\cref{app:probe}), per-arm budget-accounting rules with billed-confirmed
unit prices (\cref{app:cost}), and full run trajectories indexed by run
identifier. Two caveats are declared rather than hidden: provider-side seed
determinism is best effort, so exact token-level replay is not guaranteed
(\cref{app:cost}), and raw annotation spreadsheets live with the experiment
repository referenced in the run index.

\section*{AI Use Statement}
Large language models are the object of study; the models evaluated are
named in \cref{sec:exp:setup}. In producing the paper itself, an AI
assistant was used to draft LaTeX infrastructure and prose under close
author direction, including this statement. All experimental numbers enter
the text exclusively through a reviewed macro file generated from the
registered run reports; the wording of statistical claims is governed by a
project style rule set enforced by an automated checker; and the authors
reviewed all content and bear full responsibility for it. No model was used
to generate, select, or filter experimental results.

\bibliographystyle{style/iclr2027_conference}
\bibliography{refs}

\appendix
\numberwithin{figure}{section}
\numberwithin{table}{section}

\section{Literature Search Protocol and Wording Table}
\label{app:literature}

\paragraph{Protocol.}
Every positioning claim in the paper was made only after a neighbour search
along a fixed keyword matrix -- revocation verbs (\emph{revoke, retract,
rescind, negate, override, update instruction}) crossed with
\emph{multi-turn} and with instrument words (\emph{benchmark, evaluation,
metric}) -- run over arXiv full text, the ACL Anthology, Semantic Scholar,
and GitHub, with retrieval dates archived alongside the queries (initial
pass 2026-07-26; systematic re-run scheduled before submission). System and
dataset names were separately screened in the same four sources plus Google
Scholar; an earlier system name was abandoned after colliding with a
contract-adherence evaluation for code generation \citep{lim2025contracteval}, and
the current pair was adopted only after the name screen came back clean.

\paragraph{Near-neighbour record.}
For each borrowed idea the audit recorded the closest neighbour and one
differential sentence, which the main text inherits verbatim in spirit:
survival-style analysis of dialogue robustness \citep{li2025time}
(we keep a cross-sectional current-status reading because relapse can
remit); negative-control methodology from observational studies
\citep{ye2025role} and causal text-intervention estimation
\citep{guo2024estimating} (we systematize negative controls for black-box
clause-effect measurement and threshold calibration); contextual-bandit
selection over models \citep{poon2025online} (our arm space would be
clause-level interventions, and the H2 null demoted this to an open
question); and the multi-turn constraint-following benchmark family
\citep{canaverde2026sequor,jia2025battle,lee2026mcjudgebench} (we measure only the revoked-clause
slice, on executable checkers rather than model judges).

\paragraph{Wording table.}
The following substitutions are binding for the whole paper and are enforced
by an automated style checker over the source; the left column does not
appear in the paper (this table quotes it, under an explicit checker
exemption).

\begin{center}
\small
\begin{tabular}{p{0.42\linewidth}p{0.5\linewidth}}
  \toprule
  Forbidden & Used instead \\
  \midrule
  the first benchmark for \ldots & 
    an evaluation slice that isolates behavioral relapse of previously
    adopted, later revoked constraints \\
  binding power; a novel metric & 
    incremental behavioral effect on a single-checker projection \\
  ``significant'' for secondary outcomes &
    nominal 95\% CI, labeled \exploratory{} \\
  immediate and delayed are ``parallel'' & 
    monotone in both conditions; smaller in magnitude under immediate
    \descriptive{} \\
  dose--response (four/five arms) &
    monotone ordering \descriptive{} \\
  ``no relapse'' unqualified & 
    zero counts with denominator and rule-of-three bound (e.g.\
    \RebindZeroScaling{}) \\
  cache-deduplicated randomness & 
    best-effort provider-side seed determinism \\
  half-life; first-crossing language & 
    depth at which prevalence first crosses one half; cross-sectional
    current-status prevalence \\
  zero training, zero infrastructure & 
    no weight updates, no self-hosted GPUs \\
  discovers a new generation mechanism &
    clause-level behavioral diagnosis and control; mechanism left to
    white-box work \\
  \bottomrule
\end{tabular}
\end{center}


\section{Pre-registration, Decision Record, and Registered Deviations}
\label{app:prereg}

\paragraph{Pre-registered core.}
Each experiment carries exactly one confirmatory primary test, fixed before
data collection: the load contrast under delayed revocation for the scaling
experiment ($m{=}8$ minus $m{=}2$); the prospective AUROC at the farthest
grid point against an adequacy criterion of \AurocThreshold{}; the
\armvrblind{}$\,-\,$\armrebind{} relapse difference for equal-budget
restoration; the \armbare{}$\,-\,$\armtomb{} contrast for the tombstone
counterfactual (later joined, as a declared extension, by
\armplacebo{}$\,-\,$\armtomb{}, with the family adjusted accordingly); and
the \armadaptive{}$\,-\,$\armvr{} pass-rate difference for the ladder
comparison. Frozen alongside: the probe stopping rule ($k \in
[\ProbeKMin{}, \ProbeKMax{}]$, threshold \ProbeThreshold{}, confidence
\ProbeConf{}, zero band \ProbeZeroBand{}), the budget definition
(\CapTokens{} tokens and \MaxAttempts{} attempts per episode per arm),
task-level clustering, Benjamini--Hochberg across clause families, and the
conservative parse semantics. The temporal analysis is an analysis-layer
reading of already-collected runs and carries descriptive estimates, not a
confirmatory claim.

\paragraph{Decision-record chain.}
Every change to checkers, logging, or statistical readings required a
decision record (ADR) before taking effect. The chain, in brief:
conservative checker path made default (ADR-001); provider set fixed
(ADR-002); power-simulation calibration (ADR-003); sequential stopping rule
with paired power simulation (ADR-004); intervention and orchestrator
interface, retest gain and final-check capacity (ADR-005); multi-load slice
design (ADR-006); prospective-probe design with horizon grid (ADR-007);
equal-budget restoration design (ADR-008); scale-up to the gold task subset
(ADR-009); three-arm ladder design (ADR-010); Gate-2 scope reductions
(ADR-011); scaling-grid design and its revisions (ADR-012); current-status
prevalence reading (ADR-013); tombstone counterfactual design (ADR-014);
diagnosis gold-standard operationalization (ADR-015); and the three
post-hoc extensions below (ADR-016/017/018), each declared before its data
were collected. The full chain ships with the release.

\paragraph{Registered deviations.}
\begin{itemize}
  \item \emph{Randomness description corrected.} Cross-arm common random
    numbers were first described as cache-level deduplication; the accurate
    description is best-effort provider-side seed determinism (fresh gateway
    per arm, no cache hits; \cref{app:cost}). Imperfect determinism widens
    intervals and does not bias paired contrasts.
  \item \emph{Scope reduction beyond pre-declared levers.} A second task
    domain (SQL) was cut during confirmation (ADR-011). This was a scope
    decision outside the pre-declared reduction levers and is declared as
    such, not as pre-registered.
  \item \emph{Cross-family empty-completion artifact.} The original
    cross-family grid was unusable at scale (empty completions from
    reasoning-budget exhaustion); it was deprecated, not reused, and
    replaced by the rescue grid of ADR-018.
  \item \emph{Equal-data replay for the diagnosis gold standard.} The
    diagnosis-level comparison replays human labels through the frozen
    stopping rule on the collected sample sequences (ADR-015), which caps
    replay length at the realized sampling depth.
  \item \emph{Placebo arm added post hoc.} The tombstone design originally
    lacked an equal-length placebo arm; ADR-016 added one as a declared
    extension with cross-run pairing and a directly measured drift check,
    and the confirmatory family was Bonferroni-adjusted to two contrasts.
  \item \emph{Post-hoc threshold.} The short-completion cutoff defining
    ``empty'' is a post-hoc definition; moving the cutoff between ten and
    fifty characters relocates about ten samples and changes no conclusion.
  \item \emph{Non-graceful run terminations.} Three of the four segments of
    the scaling chain ended non-gracefully (two system restarts, one
    container outage with manual cleanup). The audit found the only source
    change across the chain was an added analysis module, with the scoring
    path untouched; three episodes recomputed offline matched the logged
    results exactly. The rescue chain lost one segment to overnight
    connectivity failure and resumed with no data loss.
  \item \emph{Extension revised in flight.} The rescue grid's repetition
    count was raised from one to three after nine episodes, before more
    than ninety-eight percent of data collection, to keep the zero-rate
    bound at least as tight as the deprecated grid's; the per-model output
    cap (\KimiMaxTokensRescue{} tokens) is a declared per-model deviation.
\end{itemize}


\section{Probe Calibration}
\label{app:probe}

\paragraph{Code as simulation.}
The stopping rule was selected by paired power simulation in which a
Bernoulli sampler is injected directly into the production probe
implementation -- the same code path that later ran against live models --
so the simulated rule family and the deployed one cannot drift apart.
Candidate rules were scored on state-classification accuracy and
non-committal rate over a grid of true $(\AC{}, \BC{})$ configurations;
the frozen rule ($k \in [\ProbeKMin{}, \ProbeKMax{}]$, threshold
\ProbeThreshold{}, confidence \ProbeConf{}, zero band \ProbeZeroBand{})
achieves worst-case state accuracy \ProbeWorstAcc{} with worst-case
non-committal rate \ProbeWorstUncertain{} at a simulated mean of
\ProbeMeanSamplesSim{} samples per arm. Sequential early stopping inflates
neither side beyond what these operating characteristics absorb, and the
induced bias on \BC{} is bounded by \ProbeBiasBound{} in absolute value.
In deployment the probe averaged \ProbeMeanSamplesObs{} samples per
diagnosis -- slower posterior convergence than the pilot anticipated, which
is itself field evidence for sequential extension over fixed small samples.

\paragraph{Executed empirical null.}
The placebo-clause protocol written into the pre-registration was executed
as a declared extension (ADR-017): \NullNClauses{} placebo clauses --
administrative sentences with no code-behavior content, matched in form and
length band to marker clauses, inserted at a fixed mid-list position into
local ledger copies -- probed under fixed paired sampling at $k = \KDiag{}$
with early stopping disabled, split between a low-adherence and a
high-adherence checker regime. The low-adherence regime returned uniformly
zero $|\hat{B}|$ (degenerate but confirming no spurious effect); the
high-adherence regime gives $p_{95}(|\hat{B}|) = \NullPNinetyFive{}$ with
maximum \NullMax{} and mean \NullMean{}. The ninety-fifth percentile sits
exactly at the zero band, and matches the pure binomial sampling-noise
prediction at this depth ($1.96\sigma = \NullTheorySigma{}$): the empirical
null shows no positional or length artifact beyond sampling noise, the zero
band equals the natural two-sigma band of the design, and no adjustment is
indicated. Two facts are stated rather than smoothed over: the boundary has
no margin, and the placebo clause occupied a single insertion position.


\section{\dataname{} Construction}
\label{app:benchmark}

\paragraph{Script anatomy.}
A slice is a scripted sequence of user turns; episodes elicit one final
implementation and contain no assistant turns. In order: the task turn
(problem statement with its base requirements); $\nmarkers$ marker turns,
each of the form ``additional requirement: define an extra helper function
named $X_i$ in the implementation (an empty implementation is fine)''; one
or more auxiliary turns with task-neutral requests (e.g.\ noting time
complexity in the docstring), which realize the delay in the delayed
condition; and, outside the control condition, a revocation turn that
withdraws the second marker by name and asks that the implementation stay
lean. Scripts are Chinese-language prompts; artifacts are Python and every
checker is language-independent. \Cref{app:examples} shows a full script
and its compiled form.

\paragraph{Marker pool and load variants.}
Marker names come from a fixed pool of five neutral identifiers, extended
to eight for the highest load; the extension is append-only and draws for
lower loads are byte-identical before and after it, locked by tests.
Markers are deliberately behavior-neutral for the task -- an empty helper
definition -- so their checkers are exact presence checks and their
satisfaction does not interact with task correctness.

\paragraph{Leakage screens.}
Two screens run over every released set. The mechanical pre-screen rejects
any auxiliary turn containing the entry-point name or any fragment of the
gold solution twenty characters or longer. The live screen prompts a model
with the auxiliary turns alone -- no task statement -- and flags leakage if
the completion passes the gold unit tests. Current sets screen clean
(\LeakMain{} main; \LeakVariants{} per variant). An early generator version
failed the live screen at \LeakVOne{} because auxiliary turns embedded
entry-point names; the generator was fixed and every released set
rescreened. The incident is reported because silent regeneration is how
leakage survives.

\paragraph{Relapse detector: attribution classes.}
Final artifacts are parsed by concatenating markdown code fences; unparseable
output is scored as non-compliant, per the conservative semantics of
\cref{sec:formulation:taxonomy}. Detected marker behavior is attributed to
one of four classes: \clstrue{} (the tombstoned clause's checker passes and
no in-force clause requires the behavior), \clsnear{} (the behavior is
independently required by an in-force clause), \clsoverlap{} (surface
overlap with a positive replacement clause), and \clslegit{} (textual
mention without the behavior). Only \clstrue{} blocks delivery and enters
relapse rates.

\paragraph{Manual audit.}
A human audit sampled \SpotCheckResult{} slices -- thirty from the main set
and twenty from each load variant -- against a fixed checklist:
well-formedness of the script, correctness of marker injection against the
pool draw, correct placement of the revocation turn, and absence of
leakage traces. Every sampled slice passed; the audit closed the obligation
registered at construction time.


\section{Human Annotation Protocols}
\label{app:annotation}

\paragraph{Annotator composition.}
Rater one is an author (the project lead); rater two is an independent
external annotator able to read Python, recruited for the dual-annotation
study, with no communication about sample judgments during annotation. All
blind ratings were collected through self-contained single-file annotation
interfaces that embed the rules, worked examples, progress saving, and CSV
export; manifests carrying checker or detector judgments were withheld,
since sending one would break blinding. Because rater one is an author, the
design leans on three safeguards: blinding at the interface level, the
independent second rater, and mechanical-evidence adjudication frozen
before annotation began.

\paragraph{Clause-mapping annotation.}
The \NClauses{} clause-to-checker mappings were annotated independently by
two annotators against the checker battery, with precision and recall
\MappingPrecision{}/\MappingRecall{} and no disagreement. One qualifier is
kept on record: the mapping instrument is template-based, and with both
raters saturated, chance-corrected agreement degenerates -- the annotation
therefore has limited power against systematic template defects. The
checker verification of \cref{sec:benchmark} (gold solutions accepted,
seeded counterexamples rejected, \CheckerVerified{}) compensates on the
executable side.

\paragraph{Blind detector audit.}
From the \DetectorStrataPool{} final artifacts of the restoration
experiment, a stratified sample of \DetectorNBlind{} was drawn under a
fixed, released seed: all \DetectorStrataFlagged{} detector-flagged
artifacts, \DetectorStrataLegit{} legitimate-reference cases, and
\DetectorStrataNone{} no-trace cases. Rater one labeled each artifact
A~(revoked behavior present), B~(reference without behavior), or C~(cannot
judge), blind to detector output and arm. Results appear in
\cref{sec:exp:validity}; the four initial disagreements were re-examined
with AST and regular-expression evidence and each proved a human miss over
a nested definition accompanied by a removal-asserting comment.

\paragraph{Diagnosis gold standard.}
For \EThreeDiagCount{} probed diagnoses collected under the frozen stopping
rule, every sampled artifact (\EThreeN{} in total) was blind-labeled on the
same A/B/C scale, and the labels were replayed through the identical
posterior classifier to produce human-side five-state diagnoses. The replay
reproduces the system's diagnoses at \EThreeDiagAgreement{} on confident
cases (\EThreeDiagN{}), with the residual disagreements driven by the same
conservative format boundary as the sample-level divergences. C-labeled
samples stay out of numerators and denominators and are listed separately.

\paragraph{Dual annotation and adjudication.}
The second-annotation protocol was frozen before rater two began: full
re-rating of both sets, three-class $\kappa$ as the primary agreement
figure, two-class $\kappa$ where both raters commit, per-rater C-usage
reported separately (\DetectorCRateOne{} versus \DetectorCRateTwo{} on the
detector set), and adjudication by mechanical evidence -- presence of the
revoked definition for the detector set; checker-relevant mechanical facts,
with format-unstable samples judged on behavioral semantics, for the
diagnosis set -- recorded append-only beside the untouched raw CSVs. On the
diagnosis set the two raters agreed on every sample
(\EThreeSecondAgreement{}); on the detector set, agreement was
\DetectorInterAgreement{} with all \DetectorDisagreements{} disagreements
at the cannot-judge boundary, each resolved by the mechanical evidence
($\kappa$ \DetectorInterKappaThree{} over three classes,
\DetectorInterKappaTwo{} where both raters commit). One qualifier governs
every adjudicated figure: the blind single-rater values are primary, because
the adjudication rule (definition-statement presence) shares semantics with
the presence checker, so post-adjudication alignment -- including the
corrected precision and the adjudicated inter-rater figures
\DetectorAdjPrecision{}/\DetectorAdjFPR{} -- is not a fully independent
validation of the detector.


\section{Additional Results}
\label{app:results}

\paragraph{Constraint-load grid, both conditions.}
At the operating point (\armbare{}), relapse rates by injected load:

\begin{center}
\small
\begin{tabular}{lccc}
  \toprule
  Condition & $m{=}2$ & $m{=}5$ & $m{=}8$ \\
  \midrule
  delayed   & \ScaleDelayedMTwo{} & \ScaleDelayedMFive{} & \ScaleDelayedMEight{} \\
  immediate & \ScaleImmediateMTwo{} & \ScaleImmediateMFive{} & \ScaleImmediateMEight{} \\
  \bottomrule
\end{tabular}
\end{center}

Both conditions are monotone in load; the immediate condition is smaller in
magnitude throughout \descriptive{}. Compiled cells are pooled at
\RebindZeroScaling{}; the ceiling control pools at \MaxZeroScaling{}.

\paragraph{Cross-family artifact history and rescue.}
The deprecated cross-family grid was dominated by empty completions from
reasoning-budget exhaustion at the shared output cap: \KimiEmptyBareOld{}
of \armbare{} episodes and \KimiEmptyRebindOld{} under compilation, rising
with load. Because empty completions cannot contain a marker definition,
relapse point estimates were not biased by the artifact -- only score and
parse metrics were -- but denominators were not answerable, and the grid
was deprecated rather than reused. The rescue pilot raised the per-model
cap to \KimiMaxTokensRescue{} tokens: \armbare{} empties fell from
\KimiPilotBareBefore{} to \KimiPilotBareAfter{}, while the compiled arm
improved only from \KimiPilotRebindBefore{} to \KimiEmptyRebindRescue{} and
stayed unusable, so the full rescue ran \armbare{} only. Clean-grid empty
rates are \KimiCleanEmptyRates{} across loads, per-cell denominators
\KimiCleanN{}, per-cell bounds \KimiCleanCellUBs{}, pooled
\KimiCleanZero{}. The short-completion cutoff is post hoc; moving it
between ten and fifty characters relocates about ten samples and changes
no conclusion.

\begin{figure}[t]
  \centering
  \includegraphics[width=0.5\linewidth]{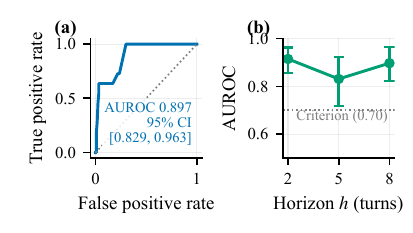}
  \caption{The probe is prospectively informative. (a)~ROC for predicting
    later relapse from the diagnosis-time probe signal at the pre-registered
    horizon $\horizon{=}8$: AUROC \AurocPrimary{} (95\% CI
    \AurocPrimaryCI{}), \NSlicesRevoked{} slices over \NTasks{} task
    clusters; the dotted diagonal is chance. (b)~AUROC across probe
    horizons (\AurocHTwo{}, \AurocHFive{}, \AurocHEight{}) with 95\%
    cluster-bootstrap CIs: prospective value stays above the pre-registered
    adequacy criterion (\AurocThreshold{}, dotted line) at every horizon,
    with no monotone trend.}
  \label{fig:diagnosis}
\end{figure}

\paragraph{Prospective AUROC, graphical view.}
\Cref{fig:diagnosis} shows the ROC at the pre-registered horizon and the
AUROC--horizon curve summarized numerically in \cref{sec:exp:validity}.

\paragraph{Ladder comparison, secondary outcomes.}
Final-attempt format instability by arm: \LadderParseFailVr{} (\armvr{}),
\LadderParseFailFixed{} (\armfixed{}), \LadderParseFailAdaptive{}
(\armadaptive{}); cluster sign test $p = \LadderParseFailP{}$,
\exploratory{} and uncorrected. Applied interventions at the chain level:
\LadderAdaptiveLOne{} \Lone{} and \LadderAdaptiveLFive{} \Lfive{}
applications under \armadaptive{}, against \LadderFixedLOne{} \Lone{}
applications under \armfixed{} -- the deliberately mismatched constant
intervention achieves a comparable format-instability drop, which blocks
attribution of the drop to structural forcing specifically. Mechanism upper
bounds: \LadderUBIntervention{} for intervening at all,
\LadderUBAdaptivity{} for adaptivity on top. All three arms show zero
observed relapse; the design carries no \armbare{} arm, so this re-states
the compiled-form result rather than re-verifying it.

\begin{figure}[t]
  \centering
  \includegraphics[width=0.5\linewidth]{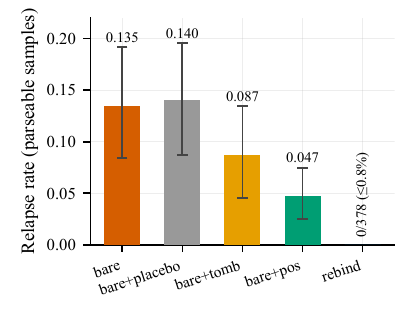}
  \caption{Counterfactual tombstones, five arms: relapse rate on parseable
    samples under \armbare{}, a near-equal-length irrelevant note
    (\armplacebo{}), the tombstone note (\armtomb{}), a positive-replacement
    note (\armpos{}), and full compilation (\armrebind{}, whose zero
    carries \TombRebindZero{}). Whiskers are 95\% cluster-bootstrap CIs;
    the ordering is monotone \descriptive{}, and no trend is implied
    between arms. The tombstone family declares two confirmatory contrasts
    (Bonferroni-adjusted): \armbare{}$\,-\,$\armtomb{} $=$ \TombPrimary{}
    \TombPrimaryCI{} and \armplacebo{}$\,-\,$\armtomb{} $=$
    \PlaceboTombDiff{} \PlaceboTombDiffCI{}. The placebo arm comes from an
    independent run paired by slice and repetition; directly re-measured
    drift is \PlaceboDriftCI{} around zero. The \armbare{}--\armplacebo{}
    difference sits within noise and implies no harm direction.}
  \label{fig:tombstone}
\end{figure}

\paragraph{Tombstone counterfactual, arm-level view.}
\Cref{fig:tombstone} shows the five arms whose contrasts are reported in
\cref{sec:exp:restoration} and \cref{tab:restoration}.

\paragraph{Tombstone score differences and worst case.}
Score differences against \armbare{}, nominal 95\% CIs, all
\exploratory{}: \armtomb{} \TombScoreNeutral{} \TombScoreNeutralCI{};
\armpos{} \TombScorePositive{} \TombScorePositiveCI{}; \armrebind{}
\TombScoreHarm{} \TombScoreHarmCI{}. Format instability under compilation
is \TombParseFailCompile{} against \TombParseFailBare{} for \armbare{}. The
worst-case bound counts every unparseable compiled episode as relapse and
still leaves the primary contrast at \TombWorstCase{} \TombWorstCaseCI{};
the direction does not flip.

\paragraph{Temporal-analysis sensitivity.}
Unparseable episodes are rare in the temporal analysis
(\SurvivalParseFail{}); recoding all of them as relapse displaces no
prevalence point by more than \SurvivalPfMaxShift{}. The anchor at $(0,1)$
is a convention, and the directly sampled at-revocation reference
(\cref{sec:exp:temporal}) is the empirical check on it; filler-turn
prefixes are nested across depths, so depth effects and prefix content are
coupled by design and read descriptively.


\section{Compiled Specifications and Intervention Examples}
\label{app:examples}

\begin{figure}[t]
  \centering
  \includegraphics[width=\linewidth]{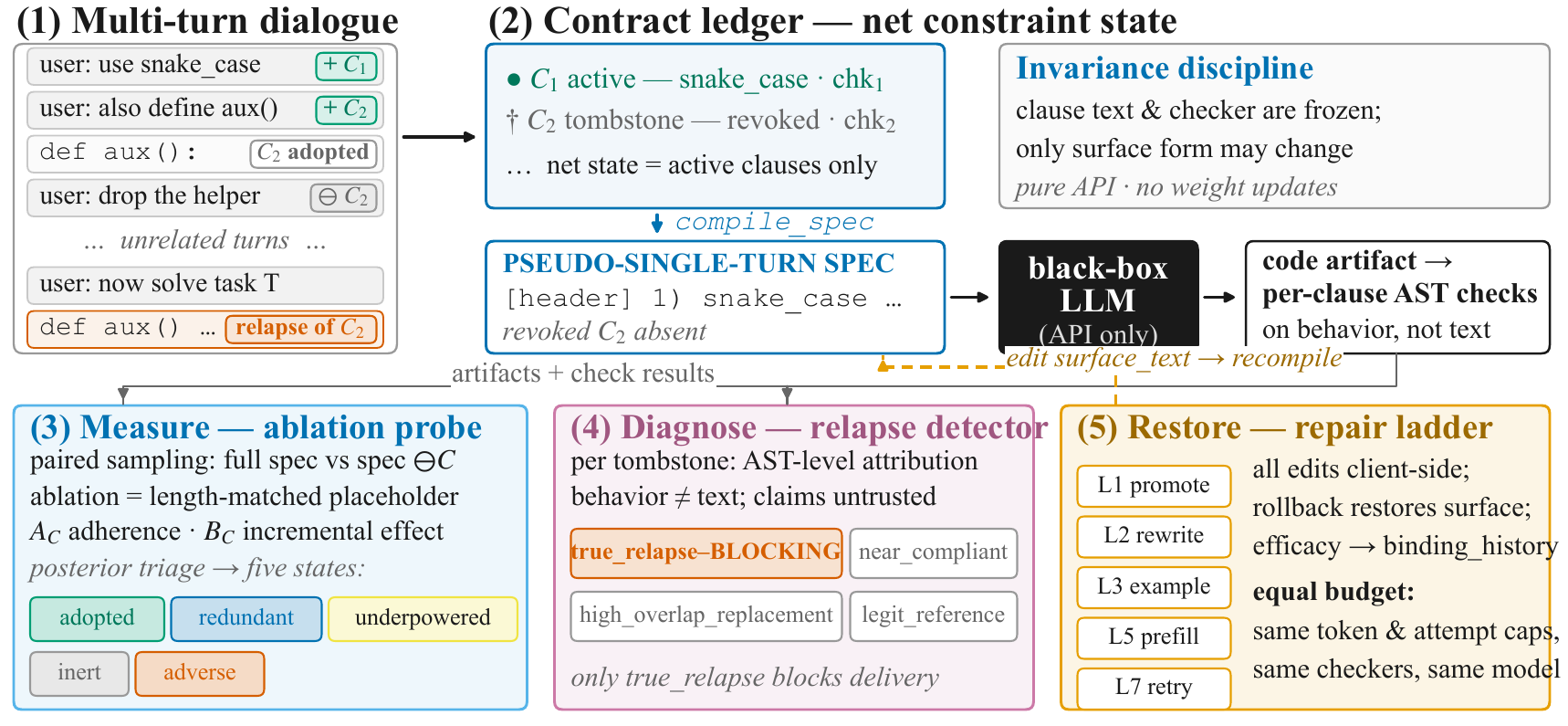}
  \caption{The \sysname{} runtime and its three pillars. Top: dialogue turns
    update a contract ledger (clauses, tombstones, binding histories); the
    ledger's net state is compiled ahead of time into a single specification;
    the model's artifact is scored by per-clause executable checkers. Bottom:
    measurement (equal-length neutral-placeholder ablation with a sequential
    probe), diagnosis (five-state triage, \cref{tab:taxonomy}), and
    restoration (repair ladder under matched budgets), with checker outcomes
    feeding back into binding histories. Interventions touch only clause
    surface text and compilation parameters; checkers are held fixed
    (\cref{sec:method:ledger}).}
  \label{fig:method}
\end{figure}

\paragraph{A script and its compilation.}
The \armbare{} arm sees the dialogue script as-is (translated; original
Chinese ships with the release), here for a palindrome task at load
$\nmarkers = \NMarkersMain$ with the second marker revoked:

\begin{quote}\small
``Implement the function below, using only the Python standard library:
[task statement for \texttt{make\_palindrome}] \\
Additional requirement: define an extra helper named \texttt{audit\_log}
(an empty implementation is fine). \\
Additional requirement: \ldots{} \texttt{sanitize\_args} \ldots \\
Additional requirement: \ldots{} \texttt{verify\_bounds} \ldots \\
Additional requirement: \ldots{} \texttt{helper\_check} \ldots \\
Additional requirement: \ldots{} \texttt{validate\_input} \ldots \\
Also, please note the implementation's time complexity in the docstring. \\
The earlier requirement about \texttt{sanitize\_args} is cancelled: do not
define \texttt{sanitize\_args} any more; keep the implementation lean.''
\end{quote}

The \armrebind{} arm sees the compiled net state instead -- a task header
and a numbered in-force clause list, with the revoked clause absent:

\begin{quote}\small
``Task: implement the function below. [task statement] \\
1.~Implement \texttt{make\_palindrome} and pass all unit tests. \\
2.~Use only the Python standard library. \\
3.--6.~Additionally define helpers \texttt{audit\_log},
\texttt{verify\_bounds}, \texttt{helper\_check}, \texttt{validate\_input}.''
\end{quote}

\paragraph{Ladder interventions in the wild.}
Intervention events are logged per attempt. A \texttt{clause\_miss} routes
to \Lone{} (the clause enters the final-check block); a \texttt{parse\_fail}
routes to \Lfive{} (the response is prefilled with an opening code fence).
The \Lseven{} feedback template -- shared by the verifier-retry baseline --
returns the report and nothing else (translated):

\begin{quote}\small
``The previous implementation has the following violations; fix them and
give the complete implementation again: \\
-- clause `\ldots' is not satisfied; \\
-- the output could not be parsed as Python code; output only complete
code.''
\end{quote}

\paragraph{Relapse, verbal-compliance form.}
From the blind detector audit (episode
\texttt{HumanEval/125:delayed:2:vr\_blind}): the artifact defines the
revoked helper directly beneath a comment asserting its removal. This is
the case type that misled human annotators.

\begin{verbatim}
    # Helper functions (empty implementations)
    def verify_bounds():
        pass

    def helper_check():
        pass  # This was removed as per the latest instruction

    def validate_input():
        pass
\end{verbatim}

\paragraph{Relapse, plain behavioral form.}
From the tombstone counterfactual (\armbare{} arm, episode
\texttt{HumanEval/102:immediate:1:bare}): the revoked helper reappears with
a docstring conceding it is not needed, and the model's closing prose
(translated) still claims ``\texttt{verify\_bounds} has been un-defined to
keep the implementation lean.''

\begin{verbatim}
    def verify_bounds():
        """Empty implementation for bounds verification
        (no longer needed)."""
        pass
\end{verbatim}

\paragraph{Frozen note texts.}
The three notes of the tombstone counterfactual, frozen in configuration
before their runs (translated; lengths are matched in the original
language, where the placebo's \PlaceboLengthChars{} characters sit inside
the instantiated tombstone notes' range):

\begin{itemize}
  \item Tombstone note: ``Reminder: the earlier requirement to
    `additionally define \{token\} in the implementation' has been revoked;
    follow the latest requirements.''
  \item Positive-replacement note: ``Reminder: \{token\} is no longer
    needed. Keep the implementation lean: apart from the task function and
    still-required helpers, do not define other helper functions.''
  \item Placebo note: ``Reminder: the project's document-numbering rules
    were recently updated; when archiving or citing materials, use the
    latest numbers.''
\end{itemize}


\section{Cost Accounting and Infrastructure}
\label{app:cost}

\paragraph{Budget metering.}
Every episode runs under identical caps in every arm: \CapTokens{} tokens
and \MaxAttempts{} attempts, metered per arm by a shared budget meter.
Ladder interventions are client-side transformations and consume no model
call to construct; retries, where a rung uses them, draw on the same
metered attempt budget. Delivery-basis figures count the tokens behind a
delivered task; the closed-loop basis adds sequential-probe tokens without
cross-task amortization.

\begin{table}[t]
  \centering
  \caption{Cost accounting. Delivery and operating-point overheads are token
    ratios and independent of unit prices; dollar figures use
    billed-confirmed unit prices. The closed-loop figure is per diagnosed
    task without cross-task amortization. Full accounting rules are in
    \cref{app:cost}.}
  \label{tab:cost}
  \small
  \setlength{\tabcolsep}{5pt}
  \begin{tabular}{lll}
    \toprule
    Quantity & Value & Basis \\
    \midrule
    Unconstrained anchor (pass@1) & \EZeroPassAtOne{} \EZeroPassAtOneCI{} &
      no clauses; \EZeroReps{} reps per task \\
    Delivery overhead & \CostDeliveryFactor{} &
      tokens per task, \armrebind{} vs \armbare{} \\
    Operating-point overhead & \CostOperatingFactor{} &
      same ratio at the operating-point measurement \\
    Closed loop with probing & \CostClosedLoopFactor{} &
      probe vs solve tokens, un-amortized \\
    API compute, main experiments & \CostMain{} &
      billed-confirmed unit prices \\
    API compute, cross-family rescue & \CostRescue{} &
      declared extension (\cref{app:prereg}) \\
    API compute, total & \CostTotal{} & sum of the two rows above \\
    \bottomrule
  \end{tabular}
\end{table}

\paragraph{Unit prices and per-model totals.}
Unit prices (USD per million tokens, input/output, cache-miss basis) were
confirmed against provider billing on 2026-08-05: \PriceEightB{} for
qwen3-8b, \PriceMax{} for qwen3-max, \PriceKimi{} for kimi-k2.7-code (the
last corrected upward from an earlier listing during the audit). Per-model
totals: \CostQwenEightB{} (qwen3-8b), \CostQwenMax{} (qwen3-max),
\CostKimi{} (kimi-k2.7-code, dominated by the rescue rerun). Project total
\CostTotal{}, reported split as \CostMain{} for the main experiments and
\CostRescue{} for the declared cross-family rescue (pilot
\CostRescuePilot{}, full chain \CostRescueFull{}); the \CostMain{}
main-experiment figure includes the deprecated cross-family grid, while
the \CostRescue{} extension covers the rescue chains only.

\paragraph{Seeds and provider-side determinism.}
Per-request seeds derive deterministically from content: a hash of the
episode key, arm, and attempt, truncated to the seed width, so arms share
randomness by construction wherever the provider honors seeds. The honoring
is best effort: gateways are created fresh per arm, no cache hits occurred
across the chains, and one identical request returned \CRNTokenExample{}
tokens on two sends -- hence ``best-effort provider-side seed
determinism,'' which widens intervals and does not bias paired contrasts.

\paragraph{Interrupted chains and their audits.}
The scaling grid ran as a four-segment chain; three segments ended
non-gracefully (two system restarts, one container outage with manual
cleanup). The audit recorded that the only source change between the first
segment and chain end was an added analysis module with the scoring path
untouched, and three episodes recomputed offline matched the logged results
exactly. The rescue grid ran as a three-segment chain, lost one segment to
overnight connectivity failure, and resumed with no data loss. All chains
resume from persisted episode state; deduplication is by episode key.

\paragraph{Recomputation entry points.}
Every reported number carries a run identifier; the run index in the
experiment repository maps identifiers to configurations, seeds, and
artifacts. This submission's release ships the reports, protocols,
generators, and trajectories; raw annotation spreadsheets and source live
with the experiment repository named in the run index.

\end{document}